\documentclass[11pt,a4paper]{article}
\usepackage[hyperref]{acl2018}
\usepackage{times}
\usepackage{latexsym}
\usepackage{booktabs}
\usepackage{array}
\usepackage{graphicx}
\usepackage{CJKutf8}
\usepackage{multirow}
\usepackage[flushleft]{threeparttable}
\usepackage{amsmath}

\usepackage{url}

\aclfinalcopy 

\title{Code-Switching Language Modeling using Syntax-Aware \\Multi-Task Learning}

\author{Genta Indra Winata, Andrea Madotto, Chien-Sheng Wu, Pascale Fung \\
        Center for Artificial Intelligence Research (CAiRE)\\
        Department of Electronic and Computer Engineering\\
        Hong Kong University of Science and Technology, Clear Water Bay, Hong Kong\\ 
        \tt \{giwinata, eeandreamad, cwuak\}@ust.hk, pascale@ece.ust.hk}

\date{}

\begin{document}
\begin{CJK*}{UTF8}{gbsn}

\maketitle
\begin{abstract}
Lack of text data has been the major issue on code-switching language modeling. In this paper, we introduce multi-task learning based language model which shares syntax representation of languages to leverage linguistic information and tackle the low resource data issue. Our model jointly learns both language modeling and Part-of-Speech tagging on code-switched utterances. In this way, the model is able to identify the location of code-switching points and improves the prediction of next word. Our approach outperforms standard LSTM based language model, with an improvement of 9.7\% and 7.4\% in perplexity on SEAME Phase I and Phase II dataset respectively. 
\end{abstract}

\section{Introduction}
Code-switching has received a lot of attention from speech and computational linguistic communities especially on how to automatically recognize text from speech and understand the structure within it. This phenomenon is very common in bilingual and multilingual communities. For decades, linguists studied this phenomenon and found that speakers switch at certain points, not randomly and obeys several constraints which point to the code-switched position in an utterance~\cite{poplack1980sometimes,belazi1994code, myers1997duelling, muysken2000bilingual, auer2007handbook}. These hypotheses have been empirically proven by observing that bilinguals tend to code-switch intra-sententially at certain (morpho)-syntactic boundaries~\cite{poplack2015code}. \citet{belazi1994code} defined the well-known theory that constraints the code-switch between a functional head and its complement is given the strong relationship between the two constituents, which corresponds to a hierarchical structure in terms of Part-of-Speech (POS) tags. \citet{muysken2000bilingual} introduced Matrix-Language Model Framework for an intra-sentential case where the primary language is called Matrix Language and the second one called Embedded Language~\cite{myers1997duelling}. A language island was then introduced which is a constituent composed entirely of the language morphemes. From the Matrix-Language Frame Model, both matrix language (ML) island and embedded language (EL) islands are well-formed in their grammars and the EL islands are constrained under ML grammar~\cite{namba2004overview}. \cite{fairchild2017determiner} studied determiner–noun switches in Spanish–English bilinguals~. 

Code-switching can be classified into two categories: intra-sentential and inter-sentential switches~\cite{poplack1980sometimes}. Intra-sentential switch defines a shift from one language to another language within an utterance. Inter-sentential switch refers to the change between two languages in a single discourse, where the switching occurs after a sentence in the first language has been completed and the next sentence starts with a new language. The example of the intra-sentential switch is shown in (1), and the inter-sentential switch is shown in (2).

\begin{itemize}
  \item[(1)] \textnormal{我 } \textnormal{要 } \textnormal{去 } check.
  \item[] \textbf{(I want to go)} check.
  \item[(2)] \textnormal{我 } \textnormal{不 } \textnormal{懂 } \textnormal{要 } \textnormal{怎么 } \textnormal{讲 } \textnormal{一 } \textnormal{个 } \textnormal{小时 } seriously I didn't have so much things to say
  \item[] \textbf{(I don’t understand how to speak for an hour)} seriously I didn’t have so much things to say
\end{itemize}

Language modeling using only word lexicons is not adequate to learn the complexity of code-switching patterns, especially in a low resource setting. Learning at the same time syntactic features such as POS tag and language identifier allows to have a shared grammatical information that constraint the next word prediction. Due to this reason, we propose a multi-task learning framework for code-switching language modeling task which is able to leverage syntactic features such as language and POS tag. 

The main contribution of this paper is two-fold. First, multi-task learning model is proposed to jointly learn language modeling task and POS sequence tagging task on code-switched utterances. Second, we incorporate language information into POS tags to create bilingual tags - it distinguishes tags between Chinese and English. The POS tag features are shared towards the language model and enrich the features to better learn where to switch. From our experiments result, we found that our method improves the perplexity on SEAME Phase I and Phase II dataset~\cite{SEAME2015}.

\section{Related Work}
The earliest language modeling research on code-switching data was applying linguistic theories on computational modelings such as Inversion Constraints and Functional Head Constraints on Chinese-English code-switching data~\cite{li2012code, ying2014language}. \citet{vu2012first} built a bilingual language model which is trained by interpolating two monolingual language models with statistical machine translation (SMT) based text generation to generate artificial code-switching text. \citet{adel2013recurrent, adel2013combination} introduced a class-based method using RNNLM for computing the posterior probability and added POS tags in the input. \citet{adel2015syntactic} explored the combination of brown word clusters, open class words, and clusters of open class word embeddings as hand-crafted features for improving the factored language model. In addition, \citet{dyer2016recurrent} proposed a generative language modeling with explicit phrase structure. A method of tying input and output embedding helped to reduce the number of parameters in language model and improved the perplexity~\cite{press2017using}.


Learning multiple NLP tasks using multi-task learning have been recently used in many domains~\cite{collobert2011natural, luong2016multi, hashimoto2016joint}. They presented a joint many-task model to handle multiple NLP tasks and share parameters with growing depth in a single end-to-end model. A work by~\citet{aguilar2017multi} showed the potential of combining POS tagging with Named-Entity Recognition task.

\section{Methodology}

This section shows how to build the features and how to train our multi-task learning language model. Multi-task learning consists of two NLP tasks: Language modeling and POS sequence tagging.

\subsection{Feature Representation}

In the model, word lexicons and syntactic features are used as input.

\textbf{Word Lexicons: } Sentences are encoded as 1-hot vectors and our vocabulary is built from training data.

\textbf{Syntactic Features: } For each \textit{language island}, phrase within the same language, we extract POS Tags iteratively using Chinese and English Penn Tree Bank Parser~\cite{tseng2005morphological, toutanova2003feature}. There are 31 English POS Tags and 34 Chinese POS Tags. Chinese words are distinguishable from English words since they have different encoding. We add language information in the POS tag label to discriminate POS tag between two languages.


\subsection{Model Description}
\begin{figure}[t]
  \centering
  \includegraphics[width=0.98\linewidth]{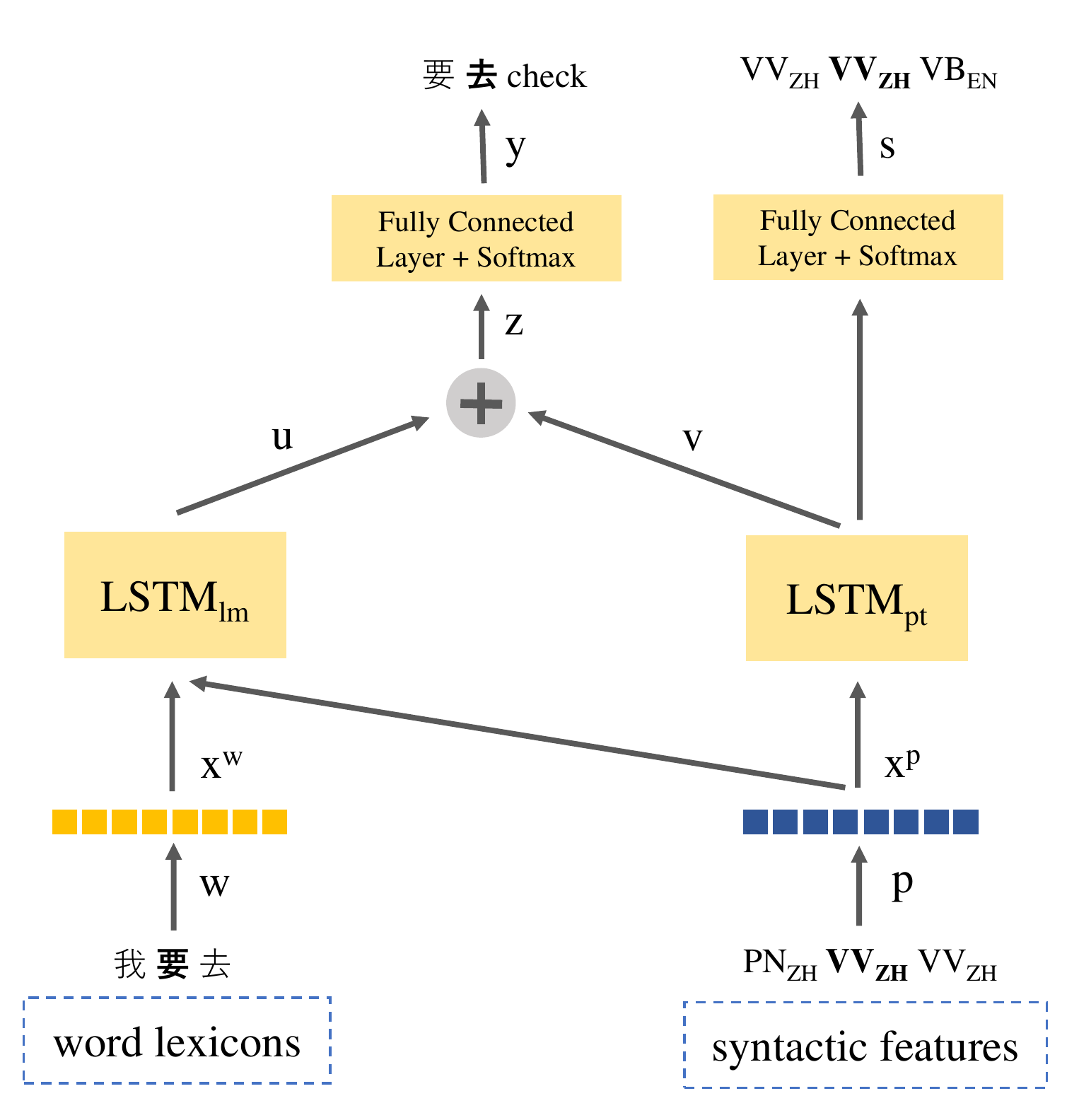}
  \caption{Multi-Task Learning Framework}
  \label{fig:multi-task-model}
\end{figure}

faFigure \ref{fig:multi-task-model} illustrates our multi-task learning extension to recurrent language model. In this multi-task learning setting, the tasks are language modeling and POS tagging. The POS tagging task shares the POS tag vector and the hidden states to LM task, but it does not receive any information from the other loss. Let $w_t$ be the word lexicon in the document and $p_t$ be the POS tag of the corresponding $w_t$ at index $t$. They are mapped into embedding matrices to get their $d$-dimensional vector representations $x_t^w$ and $x_t^p$. The input embedding weights are tied with the output weights. We concatenate $x_t^w$ and $x_t^p$ as the input of $\textnormal{LSTM}_{lm}$. The information from the POS tag sequence is shared to the language model through this step.
\[ u_t = \textnormal{LSTM}_{lm}(x_t^{w} \oplus x_t^{p}, u_{t-1}) \]
\[ v_t = \textnormal{LSTM}_{pt}(x_t^{p}, v_{t-1}) \]

where $\oplus$ denotes the concatenation operator, $u_t$ and $v_t$ are the final hidden states of $\textnormal{LSTM}_{lm}$ and $\textnormal{LSTM}_{pt}$ respectively. $u_t$ and $v_t$, the hidden states from both LSTMs are summed before predicting the next word.
\[ z_t = u_t + v_t \]
\[ y_t = \frac{e^{z_t}}{\sum_{j=1}^T e^{z_j}} \textnormal{, where } j = 1 \textnormal{, .., }T \]



The word distribution of the next word $y_{t}$ is normalized using softmax function. The model uses cross-entropy losses as error functions $\mathcal{L}_{lm}$ and $\mathcal{L}_{pt}$ for language modeling task and POS tagging task respectively. We optimize the multi-objective losses using the Back Propagation algorithm and we perform a weighted linear sum of the losses for each individual task.
\[\mathcal{L}_{total} = p \mathcal{L}_{lm} + (1-p) \mathcal{L}_{pt}\]

where $p$ is the weight of the loss in the training.

\subsection{Experimental Setup}
In this section, we present the experimental setting for this task

\textbf{Corpus: }~SEAME (South East Asia Mandarin-English), a conversational
Mandarin-English code-switching speech corpus consists of spontaneously spoken interviews and conversations \cite{SEAME2015}. Our dataset (LDC2015S04) is the most updated version of the Linguistic Data Consortium (LDC) database. However, the statistics are not identical to \citet{lyu2010analysis}. The corpus consists of two phases. In Phase I, only selected audio segments were transcribed. In Phase II, most of the audio segments were transcribed. According to the authors, it was not possible to restore the original dataset. The authors only used Phase I corpus. Few speaker ids are not in the speaker list provided by the authors~\citet{lyu2010analysis}. Therefore as a workaround, we added these ids to the train set. As our future reference, the recording lists are included in the supplementary material.

\begin{table}[!htb]
\centering
\caption{Data Statistics in SEAME Phase I}
\label{data-statistics-phase-1}
\begin{tabular}{@{}rccc@{}}
\hline
\multicolumn{1}{r|}{} & \multicolumn{1}{c|}{\textbf{Train set}} & \multicolumn{1}{c|}{\textbf{Dev set}} & \textbf{Test set}\\ \hline
\multicolumn{1}{r|}{\# Speakers} & \multicolumn{1}{c|}{139} & \multicolumn{1}{c|}{8} & 8                                                        \\ \hline
\multicolumn{1}{r|}{\# Utterances} & \multicolumn{1}{c|}{45,916} & \multicolumn{1}{c|}{1,938} & 1,228                         \\ \hline
\multicolumn{1}{r|}{\# Tokens} & \multicolumn{1}{c|}{762K} & \multicolumn{1}{c|}{31K} & 17K                        \\ \hline
\multicolumn{1}{r|}{\begin{tabular}[c]{@{}r@{}}Avg. segments \\ length\end{tabular}} & \multicolumn{1}{c|}{3.67} & \multicolumn{1}{c|}{3.68}         & 3.18          \\ \hline
\multicolumn{1}{r|}{Avg. switches}                    & \multicolumn{1}{c|}{3.60}           & \multicolumn{1}{c|}{3.47}         & 3.67          \\ \hline
\end{tabular}
\end{table}

\begin{table}[!htb]
\centering
\caption{Data Statistics in SEAME Phase II}
\label{data-statistics-phase-2}
\begin{tabular}{@{}rccc@{}}
\hline
\multicolumn{1}{l}{}                                                                & \multicolumn{1}{|c|}{\textbf{Train set}} & \multicolumn{1}{c|}{\textbf{Dev set}} & \multicolumn{1}{c}{\textbf{Test set}} \\ \hline
\multicolumn{1}{r|}{\# Speakers}                                                   & \multicolumn{1}{c|}{138}                     & \multicolumn{1}{c|}{8}                     & \multicolumn{1}{c}{8}                      \\ \hline
\multicolumn{1}{r|}{\# Utterances}                                                 & \multicolumn{1}{c|}{78,815}                  & \multicolumn{1}{c|}{4,764}                 & \multicolumn{1}{c}{3,933}                  \\ \hline
\multicolumn{1}{r|}{\# Tokens}                                                     & \multicolumn{1}{c|}{1.2M}                    & \multicolumn{1}{c|}{65K}                   & \multicolumn{1}{c}{60K}                    \\ \hline
\multicolumn{1}{r|}{\begin{tabular}[c]{@{}r@{}}Avg. segment\\ length\end{tabular}} & \multicolumn{1}{c|}{4.21}                    & \multicolumn{1}{c|}{3.59}                  & \multicolumn{1}{c}{3.99}                   \\ \hline
\multicolumn{1}{r|}{Avg. switches}                                                 & \multicolumn{1}{c|}{2.94}                    & \multicolumn{1}{c|}{3.12}                  & \multicolumn{1}{c}{3.07}                   \\ \hline
\end{tabular}
\end{table}

\begin{table}[!htb]
\centering
\caption{Code-Switching Trigger Words in SEAME Phase II}
\label{trigger-words}
\begin{tabular}{@{}lclc@{}}
\hline
\textbf{POS Tag} & \multicolumn{1}{|c}{\textbf{Freq}} & \multicolumn{1}{|c}{\textbf{POS Tag}} & \multicolumn{1}{|c}{\textbf{Freq}}\\ \hline
$\textnormal{VV}_{ZH}$ & \multicolumn{1}{|c|}{107,133} & $\textnormal{NN}_{EN}$ & \multicolumn{1}{|c}{31,031} \\ \hline
$\textnormal{AD}_{ZH}$ & \multicolumn{1}{|c|}{97,681} & $\textnormal{RB}_{EN}$ & \multicolumn{1}{|c}{12,498} \\ \hline
$\textnormal{PN}_{ZH}$ & \multicolumn{1}{|c|}{92,117} & $\textnormal{NNP}_{EN}$ & \multicolumn{1}{|c}{11,734} \\ \hline
$\textnormal{NN}_{ZH}$ & \multicolumn{1}{|c|}{45,088} & $\textnormal{JJ}_{EN}$ & \multicolumn{1}{|c}{5,040} \\ \hline
$\textnormal{VA}_{ZH}$ & \multicolumn{1}{|c|}{27,442} & $\textnormal{IN}_{EN}$ & \multicolumn{1}{|c}{4,801} \\ \hline
$\textnormal{CD}_{ZH}$ & \multicolumn{1}{|c|}{20,158} & $\textnormal{VB}_{EN}$ & \multicolumn{1}{|c}{4,703} \\ \hline
\end{tabular}
\end{table}

\textbf{Preprocessing: } First, we tokenized English and Chinese word using Stanford NLP toolkit~\cite{manning-EtAl:2014:P14-5}. Second, all hesitations and punctuations were removed except apostrophe, for examples: ``let's" and ``it's". Table \ref{data-statistics-phase-1} and Table \ref{data-statistics-phase-2} show the statistics of SEAME Phase I and II corpora. Table \ref{trigger-words} shows the most common trigger POS tag for Phase II corpus.



\textbf{Training: } The baseline model was trained using RNNLM~\cite{mikolov2011rnnlm}\footnote{downloaded from Mikolov's website http://www.fit.vutbr.cz/~imikolov/rnnlm/}. Then, we trained our LSTM models with different hidden sizes [200, 500]. All LSTMs have 2 layers and unrolled for 35 steps. The embedding size is equal to the LSTM hidden size. A dropout regularization~\cite{srivastava2014dropout} was applied to the word embedding vector and POS tag embedding vector, and to the recurrent output~\cite{gal2016theoretically} with values between [0.2, 0.4]. We used a batch size of 20 in the training. EOS tag was used to separate every sentence. We chose Stochastic Gradient Descent and started with a learning rate of 20 and if there was no improvement during the evaluation, we reduced the learning rate by a factor of 0.75. The gradient was clipped to a maximum of 0.25. For the multi-task learning, we used different loss weights hyper-parameters $p$ in the range of [0.25, 0.5, 0.75]. We tuned our model with the development set and we evaluated our best model using the test set, taking perplexity as the final evaluation metric. Where the latter was calculated by taking the exponential of the error in the negative log-form.
\[ \textnormal{PPL}(w) = e^{\mathcal{L}_{total}}\]


\section{Results}
Table \ref{results-weighted-loss-phase-1} and Table \ref{results-weighted-loss-phase-2} show the results of multi-task learning with different values of the hyper-parameter $p$. We observe that the multi-task model with $p = 0.25$ achieved the best performance. We compare our multi-task learning model against RNNLM and LSTM baselines. The baselines correspond to recurrent neural networks that are trained with word lexicons. Table \ref{results-phase-1} and Table \ref{results-phase-2} present the overall results from different models. The multi-task model performs better than LSTM baseline by 9.7\% perplexity in Phase I and 7.4\% perplexity in Phase II. The performance of our model in Phase II is also better than the RNNLM (8.9\%) and far better than the one presented in~\citet{adel2013combination} in Phase I. 

\begin{figure*}[!htb]
  \centering
  \includegraphics[width=0.9\linewidth]{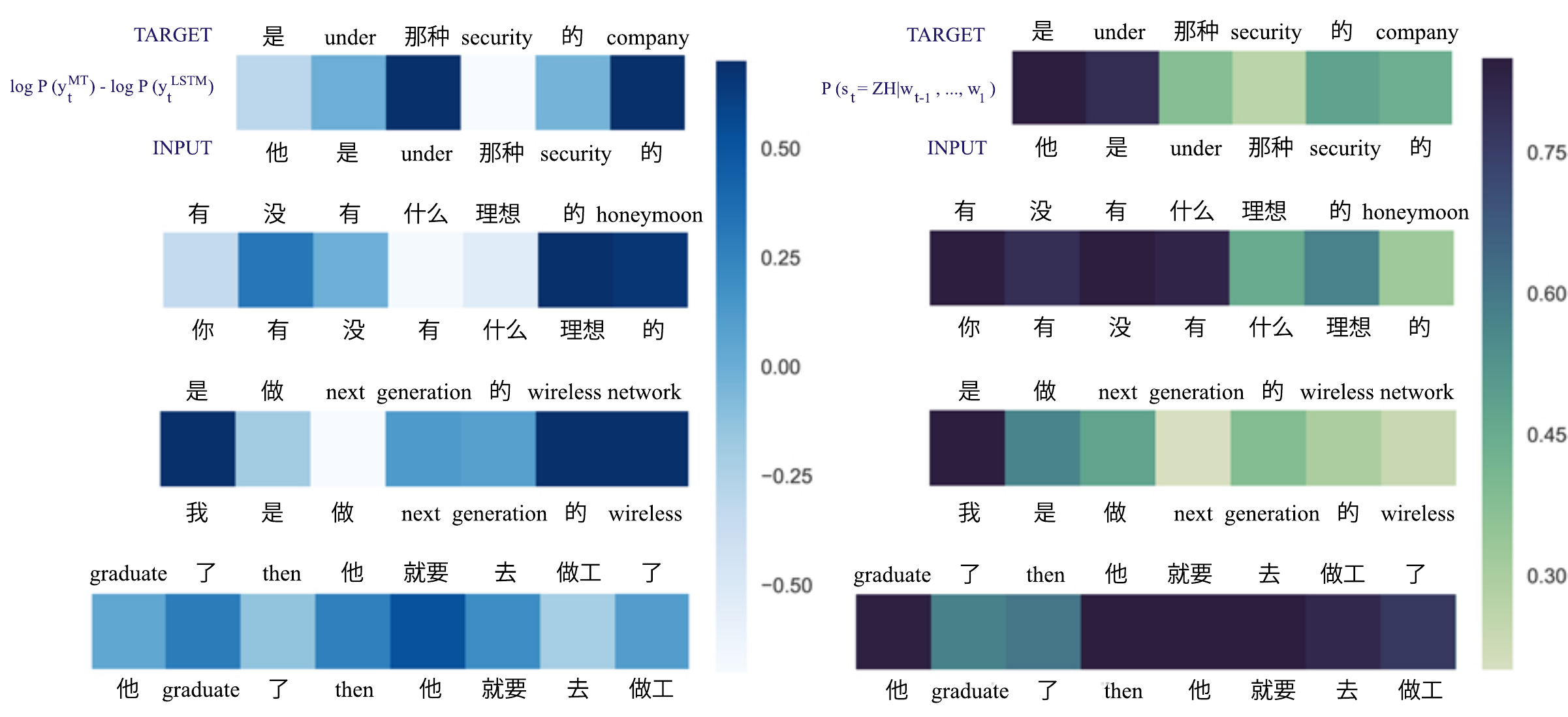}
  \caption{Prediction examples in Phase II. \textbf{Left:} Each square shows the target word's log probability improvement by multi-task model compared to LSTM model (Darker color is better).
  \textbf{Right:} Each square shows the probability of the next POS tag is Chinese (Darker color represents higher probability)}
  \label{fig:language}
\end{figure*}

\begin{table}[!tb]
\centering
\caption{Multi-task results with different weighted loss hyper-parameter in Phase I}
\label{results-weighted-loss-phase-1}
\begin{tabular}{@{}cccc@{}}
\hline
\textbf{\begin{tabular}[c]{@{}c@{}}Hidden\\ size\end{tabular}}  & \multicolumn{1}{|c|}{\textbf{$p$}} & \multicolumn{1}{c|}{\textbf{\begin{tabular}[c]{@{}c@{}}PPL\\ Dev\end{tabular}}} & \multicolumn{1}{c}{\textbf{\begin{tabular}[c]{@{}c@{}}PPL\\ Test\end{tabular}}} \\ \hline
\multirow{3}{*}{200} & \multicolumn{1}{|c|}{\textbf{0.25}} & \multicolumn{1}{c|}{\textbf{180.90}}	& \multicolumn{1}{c}{\textbf{178.18}} \\ \cline{2-4}
 & \multicolumn{1}{|c|}{0.5} & \multicolumn{1}{c|}{182.6} & \multicolumn{1}{c}{178.75}\\ \cline{2-4}
 & \multicolumn{1}{|c|}{\textbf{0.75}} & \multicolumn{1}{c|}{\textbf{180.90}} & \multicolumn{1}{c}{\textbf{178.18}} \\ \hline
\multirow{3}{*}{500} & \multicolumn{1}{|c|}{\textbf{0.25}} & \multicolumn{1}{c|}{\textbf{173.55}} & \multicolumn{1}{c}{\textbf{174.96}} \\ \cline{2-4}
 & \multicolumn{1}{|c|}{0.5} & \multicolumn{1}{c|}{175.23} & \multicolumn{1}{c}{173.89} \\ \cline{2-4}
 & \multicolumn{1}{|c|}{0.75} & \multicolumn{1}{c|}{185.83} & \multicolumn{1}{c}{178.49} \\ \hline
\end{tabular}
\end{table}

\begin{table}[!tb]
\centering
\caption{Multi-task results with different weighted loss hyper-parameter in Phase II}
\label{results-weighted-loss-phase-2}
\begin{tabular}{@{}cccc@{}}
\hline
\textbf{\begin{tabular}[c]{@{}c@{}}Hidden\\ size\end{tabular}}  & \multicolumn{1}{|c|}{\textbf{$p$}} & \multicolumn{1}{c|}{\textbf{\begin{tabular}[c]{@{}c@{}}PPL\\ Dev\end{tabular}}} & \multicolumn{1}{c}{\textbf{\begin{tabular}[c]{@{}c@{}}PPL\\ Test\end{tabular}}} \\ \hline
\multirow{3}{*}{200} & \multicolumn{1}{|c|}{\textbf{0.25}} & \multicolumn{1}{c|}{\textbf{149.68}}	& \multicolumn{1}{c}{\textbf{149.84}} \\ \cline{2-4}
 & \multicolumn{1}{|c|}{0.5} & \multicolumn{1}{c|}{150.92} & \multicolumn{1}{c}{152.38} \\ \cline{2-4}
 & \multicolumn{1}{|c|}{0.75} & \multicolumn{1}{c|}{150.32} & \multicolumn{1}{c}{151.22} \\ \hline
\multirow{3}{*}{500} & \multicolumn{1}{|c|}{\textbf{0.25}} & \multicolumn{1}{c|}{\textbf{141.86}} & \multicolumn{1}{c}{\textbf{141.71}} \\ \cline{2-4}
 & \multicolumn{1}{|c|}{0.5} & \multicolumn{1}{c|}{144.18} & \multicolumn{1}{c}{144.27} \\ \cline{2-4}
 & \multicolumn{1}{|c|}{0.75} & \multicolumn{1}{c|}{145.08} & \multicolumn{1}{c}{144.85} \\ \hline
\end{tabular}
\end{table}

\begin{table}[!tb]
\centering
\caption{Results in Phase I}
\label{results-phase-1}
\begin{tabular}{@{}lcc@{}}
\hline
\textbf{Model} & \multicolumn{1}{|c|}{\textbf{\begin{tabular}[c]{@{}c@{}}PPL \\ Dev\end{tabular}}} & \multicolumn{1}{c}{\textbf{\begin{tabular}[c]{@{}c@{}}PPL\\ Test\end{tabular}}} \\ \hline
RNNLM \cite{adel2013recurrent} & \multicolumn{1}{|c|}{246.60} & \multicolumn{1}{c}{287.88}\\ \hline
\cite{adel2015syntactic} & \multicolumn{1}{|c|}{238.86} & \multicolumn{1}{c}{245.40}\\ \hline
FI + OF \cite{adel2013recurrent} & \multicolumn{1}{|c|}{219.85} & \multicolumn{1}{c}{239.21}\\ \hline
FLM \cite{adel2013combination} & \multicolumn{1}{|c|}{177.79} & \multicolumn{1}{c}{192.08}\\ \hline
LSTM & \multicolumn{1}{|c|}{190.33} & \multicolumn{1}{c}{185.91} \\ \hline
\textnormal{+ syntactic features} & \multicolumn{1}{|c|}{178.51} & \multicolumn{1}{c}{176.57} \\ \hline
\textbf{Multi-task} & \multicolumn{1}{|c|}{\textbf{173.55}} & \multicolumn{1}{c}{\textbf{174.96}} \\ \hline
\end{tabular}
\end{table}

\begin{table}[!tb]
\centering
\caption{Results in Phase II}
\label{results-phase-2}
\begin{tabular}{@{}lcc@{}}
\hline
\textbf{Model} & \multicolumn{1}{|c|}{\textbf{\begin{tabular}[c]{@{}c@{}}PPL \\ Dev\end{tabular}}} & \multicolumn{1}{c}{\textbf{\begin{tabular}[c]{@{}c@{}}PPL\\ Test\end{tabular}}} \\ \hline
RNNLM & \multicolumn{1}{|c|}{178.35} & 171.27 \\ \hline
LSTM & \multicolumn{1}{|c|}{150.65} & 153.06 \\ \hline
\textnormal{+ syntactic features} & \multicolumn{1}{|c|}{147.44} & 148.38 \\ \hline
\textbf{Multi-task} & \multicolumn{1}{|c|}{\textbf{141.86}} & \textbf{141.71} \\ \hline
\end{tabular}
\end{table}

Moreover, the results show that adding shared POS tag representation to $\textnormal{LSTM}_{lm}$ does not hurt the performance of the language modeling task. This implies that the syntactic information helps the model to better predict the next word in the sequence. To further verify this hypothesis, we conduct two analysis by visualizing our prediction examples in Figure~\ref{fig:language}: 
\paragraph{a)}\noindent Measure the improvement of the target word's log probability by multi-task model compared to standard LSTM model. This is computed by calculating the log probability difference between two models. According to Figure~\ref{fig:language}, in most of the cases, the multi-task model improves the prediction of the monolingual segments and particularly in code-switching points such as ``under", ``security", ``generation'', ``then", ``graduate", ``他", and ``的". It also shows that the multi-task model is more precise in learning where to switch language. On the other hand, Table \ref{trigger-words} shows the relative frequency of the trigger POS tag. The word ``then" belong to $\textnormal{RB}_{EN}$, which is one of the most common trigger words in the list. Furthermore, the target word prediction is significantly improved in most of the trigger words.  
\paragraph{b)}\noindent Report the probability that the next produced POS tag is Chinese. It is shown that words ``then", ``security", ``了'', ``那种'', ``做'', and ``的''  tend to switch the language context within the utterance. However, it is very hard to predict all the cases correctly. This is may due to the fact that without any switching, the model still creates a correct sentence.

\section{Conclusion}
In this paper, we propose a multi-task learning approach for code-switched language modeling. The multi-task learning models achieve the best performance and outperform LSTM baseline with 9.7\% and 7.4\% improvement in perplexity for Phase I and Phase II SEAME corpus respectively. This implies that by training two different NLP tasks together the model can correctly learn the correlation between them. Indeed, the syntactic information helps the model to be aware of code-switching points and it improves the performance over the language model. Finally, we conclude that multi-task learning has good potential on code-switching language modeling research and there are still rooms for improvements, especially by adding more language pairs and corpora.




\section*{Acknowledgments}
This work is partially funded by ITS/319/16FP of the Innovation Technology Commission, HKUST 16214415 \& 16248016 of Hong Kong Research Grants Council, and RDC 1718050-0 of EMOS.AI.

\bibliography{acl2018}
\bibliographystyle{acl_natbib}

\end{CJK*}

\section*{Supplementary Materials}
\subsection*{Results}
Results with different hyper-parameter settings

\begin{table}[!htb]
\centering
\caption{Results in Phase I}
\label{appendix-results}
\begin{tabular}{@{}lcccccc@{}}
\hline
\textbf{Model} & \multicolumn{1}{|c|}{\textbf{\begin{tabular}[c]{@{}c@{}}Hidden\\ size\end{tabular}}} & \textbf{\begin{tabular}[c]{@{}c@{}}Embedding\\ size\end{tabular}} & \multicolumn{1}{|c|}{\textbf{Dropout}} & \textbf{POS tag dropout} & \multicolumn{1}{|c|}{\textbf{\begin{tabular}[c]{@{}c@{}}PPL \\ dev\end{tabular}}} & \textbf{\begin{tabular}[c]{@{}c@{}}PPL\\ test\end{tabular}} \\ \hline
\multirow{2}{*}{LSTM} & \multicolumn{1}{|c|}{200} & 200 & \multicolumn{1}{|c|}{0.2} & - & \multicolumn{1}{|c|}{197.5} & 196.84 \\ \cline{2-7} 
& \multicolumn{1}{|c|}{500} & 500 & \multicolumn{1}{|c|}{0.4} & - & \multicolumn{1}{|c|}{190.33} & 185.91 \\ \hline
\multirow{2}{*}{+ syntactic features} & \multicolumn{1}{|c|}{200} & 200 & \multicolumn{1}{|c|}{0.2} & - & \multicolumn{1}{|c|}{187.37} & 184.87 \\ \cline{2-7} 
& \multicolumn{1}{|c|}{500} & 500 & \multicolumn{1}{|c|}{0.4} & - & \multicolumn{1}{|c|}{178.51} & 176.57 \\ \hline
\multirow{2}{*}{Multi-task ($p=0.25$)} & \multicolumn{1}{|c|}{200} & 200 & \multicolumn{1}{|c|}{0.4} & 0.2 & \multicolumn{1}{|c|}{180.91} & 178.18 \\ \cline{2-7} 
& \multicolumn{1}{|c|}{\textbf{500}} & \textbf{500} & \multicolumn{1}{|c|}{\textbf{0.4}} & \textbf{0.4} & \multicolumn{1}{|c|}{\textbf{173.55}} & \textbf{174.96} \\ \hline
\multirow{2}{*}{Multi-task ($p=0.50$)} & \multicolumn{1}{|c|}{200} & 200 & \multicolumn{1}{|c|}{0.4} & 0.2 & \multicolumn{1}{|c|}{182.6} & 178.75 \\ \cline{2-7} 
& \multicolumn{1}{|c|}{500} & 500 & \multicolumn{1}{|c|}{0.4} & 0.4 & \multicolumn{1}{|c|}{175.23} & 173.89 \\ \hline 
\multirow{2}{*}{Multi-task ($p=0.75$)} & \multicolumn{1}{|c|}{200} & 200 & \multicolumn{1}{|c|}{0.4} & 0.2 & \multicolumn{1}{|c|}{180.90} & 178.18 \\ \cline{2-7} 
& \multicolumn{1}{|c|}{500} & 500 & \multicolumn{1}{|c|}{0.4} & 0.4 & \multicolumn{1}{|c|}{185.83} & 178.49 \\ \hline
\end{tabular}
\end{table}

\begin{table}[!htb]
\centering
\caption{Results in Phase II}
\label{appendix-results}
\begin{tabular}{@{}lcccccc@{}}
\hline
\textbf{Model} & \multicolumn{1}{|c|}{\textbf{\begin{tabular}[c]{@{}c@{}}Hidden\\ size\end{tabular}}} & \textbf{\begin{tabular}[c]{@{}c@{}}Embedding\\ size\end{tabular}} & \multicolumn{1}{|c|}{\textbf{Dropout}} & \textbf{POS tag dropout} & \multicolumn{1}{|c|}{\textbf{\begin{tabular}[c]{@{}c@{}}PPL \\ dev\end{tabular}}} & \textbf{\begin{tabular}[c]{@{}c@{}}PPL\\ test\end{tabular}} \\ \hline
\multirow{2}{*}{RNNLM} & \multicolumn{1}{|c|}{200} & 200 & \multicolumn{1}{|c|}{-} & - & \multicolumn{1}{|c|}{181.87} & 176.80 \\ \cline{2-7} 
& \multicolumn{1}{|c|}{500} & 500 & \multicolumn{1}{|c|}{-} & - & \multicolumn{1}{|c|}{178.35} & 171.27 \\ \hline 
\multirow{2}{*}{LSTM} & \multicolumn{1}{|c|}{200} & 200 & \multicolumn{1}{|c|}{0.2} & - & \multicolumn{1}{|c|}{156.77} & 159.58 \\ \cline{2-7} 
& \multicolumn{1}{|c|}{500} & 500 & \multicolumn{1}{|c|}{0.4} & - & \multicolumn{1}{|c|}{150.65} & 153.06 \\ \hline
\multirow{2}{*}{+ syntactic features} & \multicolumn{1}{|c|}{200} & 200 & \multicolumn{1}{|c|}{0.2} & - & \multicolumn{1}{|c|}{153.6} & 152.66 \\ \cline{2-7} 
& \multicolumn{1}{|c|}{500} & 500 & \multicolumn{1}{|c|}{0.4} & - & \multicolumn{1}{|c|}{147.44} & 148.38 \\ \hline
\multirow{2}{*}{Multi-task ($p=0.25$)} & \multicolumn{1}{|c|}{200} & 200 & \multicolumn{1}{|c|}{0.4} & 0.2 & \multicolumn{1}{|c|}{149.68} & 149.84 \\ \cline{2-7} 
& \multicolumn{1}{|c|}{\textbf{500}} & \textbf{500} & \multicolumn{1}{|c|}{\textbf{0.4}} & \textbf{0.4} & \multicolumn{1}{|c|}{\textbf{141.86}} & \textbf{141.71} \\ \hline
\multirow{2}{*}{Multi-task ($p=0.50$)} & \multicolumn{1}{|c|}{200} & 200 & \multicolumn{1}{|c|}{0.4} & 0.2 & \multicolumn{1}{|c|}{150.92} & 152.38 \\ \cline{2-7} 
& \multicolumn{1}{|c|}{500} & 500 & \multicolumn{1}{|c|}{0.4} & 0.4 & \multicolumn{1}{|c|}{144.18} & 144.27 \\ \hline
\multirow{2}{*}{Multi-task ($p=0.75$)} & \multicolumn{1}{|c|}{200} & 200 & \multicolumn{1}{|c|}{0.4} & 0.2 & \multicolumn{1}{|c|}{150.32} & 151.22 \\ \cline{2-7} 
& \multicolumn{1}{|c|}{500} & 500 & \multicolumn{1}{|c|}{0.4} & 0.4 & \multicolumn{1}{|c|}{145.08} & 144.85 \\ \hline
\end{tabular}
\end{table}

\clearpage
\subsection*{Recording Lists}
We split the recording ids into train, development, and test set as the following:

\begin{table}[!htb]
\centering
\caption{Recording distribution in Phase I}
\label{recording-list-phase2}
\begin{tabular}{@{}ccc@{}}
\hline
\multirow{2}{*}{\textbf{Data}} & \multicolumn{2}{|c}{\textbf{Recording list}}                                                                                                                                                                                                                                                                                                                                                                                                                                                                                                                                                                                                                                                                                                                                                                                                                                                                                                                                                                                                                                                                                                                                                                                                                                                                                                                                                                                                                                                                                                                                                                                                                                                                                                                                                                                                          \\ \cline{2-3} 
                           & \multicolumn{1}{|c|}{\textbf{Conversation}}                                                                                                                                                                                                                                                                                                                                                                                                                                                                                                                                                                                                                                                                                                                                                                                                                                                     & \textbf{Interview}                                                                                                                                                                                                                                                                                                                                                                                                                                                                                                                                                                                                                                                                                                                                                                                                                                                                                                                                                        \\ \hline
\textbf{Train}             & \multicolumn{1}{|c|}{\begin{tabular}[c]{@{}c@{}}
02NC03FBX, 02NC04FBY, 03NC05FAX \\
03NC06FAY,04NC07FBX,04NC08FBY\\
06NC11MAX,06NC12MAY,07NC13MBP\\
07NC14FBQ,08NC15MBP,08NC16FBQ\\
09NC17FBP,09NC18MBQ,10NC19MBP\\
10NC20MBQ,11NC21FBP,11NC22MBQ\\
12NC23FBP,12NC24FBQ,13NC25MBP\\
13NC26MBQ,14NC27MBP,14NC28MBQ\\
16NC31FBP,16NC32FBQ,18NC35FBP\\
18NC36MBQ,19NC37MBP,19NC38FBQ\\
21NC41MBP,22NC43FBP,22NC44MBQ\\
23NC35FBQ,23NC45MBP,24NC35FBQ\\
24NC45MBP,25NC43FBQ,25NC47MBP\\
26NC48FBP,26NC49FBQ,27NC47MBQ\\
27NC50FBP,28NC51MBP,28NC52FBQ\\
29NC53MBP,29NC54FBQ,30NC48FBP\\
30NC49FBQ,31NC35FBQ,31NC50XFB\\
32NC36MBQ,32NC50FBP,33NC37MBP\\
33NC43FBQ,34NC37MBP,35NC56MBP\\
36NC46FBQ,37NC45MBP,38NC50FBP\\
39NC57FBX,40NC58FAY,41NC59MAX\\
42NC60FBQ,44NC44MBQ,45NC22MBQ\\
46NC41MBP,46NC41MBP\\
\end{tabular}} & \begin{tabular}[c]{@{}c@{}}
NI02FAX,NI04FBX,NI05MBQ\\
NI06FBP,NI07FBQ,NI08FBP\\
NI09FBP,NI10FBP,NI11FBP\\
NI12MAP,NI13MBQ,NI14MBP\\
NI15FBQ,NI16FBP,NI17FBQ\\
NI18MBP,NI19MBQ,NI20MBP\\
NI21MBQ,NI22FBP,NI23FBQ\\
NI24MBP,NI25MBQ,NI26FBP\\
NI27MBQ,NI28MBP,NI29MBP\\
NI30MBQ,NI31FBP,NI32FBQ\\
NI33MBP,NI34FBQ,NI35FBP\\
NI36MBQ,NI37MBP,NI39FBP\\
NI40FBQ,NI41MBP,NI42FBQ\\
NI43FBP,NI44MBQ,NI45FBP\\
NI46FBQ,NI47MBP,NI48FBQ\\
NI49MBP,NI50FBQ,NI51MBP\\
NI52MBQ,NI53FBP,NI54FBQ\\
NI55FBP,NI56MBX,NI57FBQ\\
NI58FBP,NI59FBQ,NI60MBP\\
NI61FBP,NI62MBQ,NI63MBP\\
NI64FBQ,NI65MBP,NI66MBQ\\
NI67MBQ,UI02FAZ,UI03FAZ\\
UI04FAZ,UI05MAZ,UI06MAZ\\
UI07FAZ,UI08MAZ,UI10FAZ\\
UI11FAZ,UI12FAZ,UI13FAZ\\
UI14MAZ,UI15FAZ,UI16MAZ\\
UI17FAZ,UI18MAZ,UI19MAZ\\
UI20MAZ,UI21MAZ,UI22MAZ\\
UI23FAZ,UI24MAZ,UI25FAZ\\
UI26MAZ,UI27FAZ,UI28FAZ\\
UI29FAZ\\
\end{tabular} \\ \hline
\textbf{Dev}               & \multicolumn{1}{|c|}{\begin{tabular}[c]{@{}c@{}}01NC01FBX, 01NC02FBY, 15NC29FBP\\ 15NC30MBQ, 21NC42MBQ, 43NC61FBQ\end{tabular}}                                                                                                                                                                                                                                                                                                                                                                                                                                                                                                                                                                                                                                                                                                                                                            & \begin{tabular}[c]{@{}c@{}}UI01FAZ, UI09MAZ\end{tabular}                                                                                                                                                                                                                                                                                                                                                                                                                                                                                                                                                                                                                                                                                                                                                                                                                                                                                                                 \\ \hline
\textbf{Test}              & \multicolumn{1}{|c|}{\begin{tabular}[c]{@{}c@{}}05NC09FAX, 05NC10MAY, 17NC33FBP\\ 17NC34FBQ, 20NC39MBP, 20NC40FBQ\end{tabular}}                                                                                                                                                                                                                                                                                                                                                                                                                                                                                                                                                                                                                                                                                                                                                             & \begin{tabular}[c]{@{}c@{}}NI01MAX, NI03FBX\end{tabular}                                                                                                                                                                                                                                                                                                                                                                                                                                                                                                                                                                                                                                                                                                                                                                                                                                                                                                             \\ \hline
\end{tabular}
\end{table}

\clearpage
\begin{table}[!htb]
\centering
\caption{Recording distribution in Phase II}
\label{recording-list-phase2}
\begin{tabular}{@{}ccc@{}}
\hline
\multirow{2}{*}{\textbf{Data}} & \multicolumn{2}{|c}{\textbf{Recording list}}                                                                                                                                                                                                                                                                                                                                                                                                                                                                                                                                                                                                                                                                                                                                                                                                                                                                                                                                                                                                                                                                                                                                                                                                                                                                                                                                                                                                                                                                                                                                                                                                                                                                                                                                                                                                          \\ \cline{2-3} 
                           & \multicolumn{1}{|c|}{\textbf{Conversation}}                                                                                                                                                                                                                                                                                                                                                                                                                                                                                                                                                                                                                                                                                                                                                                                                                                                     & \textbf{Interview}                                                                                                                                                                                                                                                                                                                                                                                                                                                                                                                                                                                                                                                                                                                                                                                                                                                                                                                                                        \\ \hline
\textbf{Train}             & \multicolumn{1}{|c|}{\begin{tabular}[c]{@{}c@{}}
02NC03FBX, 02NC04FBY, 03NC05FAX\\ 
03NC06FAY, 04NC07FBX, 04NC08FBY\\ 
06NC11MAX, 06NC12MAY, 07NC13MBP\\ 
07NC14FBQ, 08NC15MBP, 08NC16FBQ \\ 
09NC17FBP, 09NC18MBQ, 10NC19MBP\\ 
10NC20MBQ, 11NC21FBP, 11NC22MBQ \\ 
12NC23FBP, 12NC24FBQ, 13NC25MBP\\ 
13NC26MBQ, 14NC27MBP, 14NC28MBQ \\ 
16NC31FBP, 16NC32FBQ, 18NC35FBP\\ 
18NC36MBQ, 19NC37MBP, 19NC38FBQ \\ 
21NC41MBP, 22NC43FBP, 22NC44MBQ\\ 
23NC35FBQ, 23NC45MBP, 24NC35FBQ \\ 
24NC45MBP, 25NC43FBQ, 25NC47MBP\\ 
26NC48FBP, 26NC49FBQ, 27NC47MBQ \\ 
27NC50FBP, 28NC51MBP, 28NC52FBQ\\ 
29NC53MBP, 29NC54FBQ, 30NC48FBP \\ 
30NC49FBQ, 31NC35FBQ, 31NC50XFB\\ 
32NC36MBQ, 32NC50FBP, 33NC37MBP \\ 
33NC43FBQ, 34NC37MBP, 35NC56MBP\\ 
36NC46FBQ, 37NC45MBP, 38NC50FBP \\ 
39NC57FBX, 40NC58FAY, 41NC59MAX\\ 
42NC60FBQ, 44NC44MBQ, 45NC22MBQ \\ 
46NC41MBP\end{tabular}} & \begin{tabular}[c]{@{}c@{}}
NI02FAX, NI04FBX, NI05MBQ\\ 
NI06FBP, NI07FBQ, NI08FBP\\ 
NI09FBP, NI10FBP, NI12MAP\\ 
NI13MBQ, NI14MBP, NI15FBQ\\ 
NI16FBP, NI17FBQ, NI18MBP\\ 
NI19MBQ, NI20MBP, NI21MBQ\\ 
NI22FBP, NI23FBQ, NI24MBP\\ 
NI25MBQ, NI26FBP, NI27MBQ\\ 
NI28MBP, NI29MBP, NI30MBQ\\ 
NI31FBP, NI32FBQ, NI33MBP\\ 
NI34FBQ, NI35FBP, NI36MBQ\\
NI37MBP, NI39FBP, NI40FBQ\\
NI41MBP, NI42FBQ, NI43FBP\\
NI44MBQ, NI45FBP, NI46FBQ\\
NI47MBP, NI48FBQ, NI49MBP\\ 
NI50FBQ, NI51MBP, NI52MBQ\\
NI53FBP, NI54FBQ, NI55FBP\\
NI56MBX, NI57FBQ, NI58FBP\\
NI59FBQ, NI60MBP, NI61FBP\\
NI62MBQ, NI63MBP, NI64FBQ\\
NI65MBP, NI66MBQ, NI67MBQ\\ 
UI02FAZ, UI03FAZ, UI04FAZ\\ 
UI05MAZ, UI06MAZ, UI07FAZ\\ 
UI08MAZ, UI10FAZ, UI11FAZ\\
UI12FAZ, UI13FAZ, UI14MAZ\\
UI15FAZ, UI16MAZ, UI17FAZ\\
UI18MAZ, UI19MAZ, UI20MAZ\\
UI21MAZ, UI22MAZ, UI23FAZ\\
UI24MAZ, UI25FAZ, UI26MAZ\\
UI27FAZ, UI28FAZ, UI29FAZ\end{tabular} \\ \hline
\textbf{Dev}               & \multicolumn{1}{|c|}{\begin{tabular}[c]{@{}c@{}}01NC01FBX, 01NC02FBY, 15NC29FBP\\ 15NC30MBQ, 21NC42MBQ, 43NC61FBQ\end{tabular}}                                                                                                                                                                                                                                                                                                                                                                                                                                                                                                                                                                                                                                                                                                                                                            & \begin{tabular}[c]{@{}c@{}}UI01FAZ, UI09MAZ\end{tabular}                                                                                                                                                                                                                                                                                                                                                                                                                                                                                                                                                                                                                                                                                                                                                                                                                                                                                                                 \\ \hline
\textbf{Test}              & \multicolumn{1}{|c|}{\begin{tabular}[c]{@{}c@{}}05NC09FAX, 05NC10MAY, 17NC33FBP\\ 17NC34FBQ, 20NC39MBP, 20NC40FBQ\end{tabular}}                                                                                                                                                                                                                                                                                                                                                                                                                                                                                                                                                                                                                                                                                                                                                             & \begin{tabular}[c]{@{}c@{}}NI01MAX, NI03FBX\end{tabular}                                                                                                                                                                                                                                                                                                                                                                                                                                                                                                                                                                                                                                                                                                                                                                                                                                                                                                                 \\ \hline
\end{tabular}
\end{table}

\end{document}